%% file: main.tex
\documentclass{article}
\usepackage{iclr2027_conference,times}
\input{math_commands.tex}

\usepackage{hyperref}
\usepackage{url}
\usepackage{graphicx}
\usepackage{booktabs}
\usepackage{balance}
\usepackage{color}
\usepackage{graphics} 
\usepackage{amsmath,amssymb,amsfonts} 
\usepackage{bm}
\usepackage{subcaption}	 
\usepackage{multirow}
\usepackage{array}
\usepackage{hyperref}
\usepackage{xcolor}
\usepackage{float}
\usepackage{booktabs}
\usepackage{multirow}
\usepackage{textcomp}
\usepackage{setspace}
\usepackage{tabularx}
\usepackage{makecell}
\usepackage{colortbl}  
\usepackage{pifont}
\usepackage{wrapfig}
\usepackage{titletoc}
\usepackage{placeins}

\newcolumntype{Y}{>{\centering\arraybackslash}X}
\newcolumntype{C}[1]{>{\centering\arraybackslash}p{#1}}
\newcommand{\methodname}{VEOcc++}

\hypersetup{
    hidelinks
}

\title{\methodname{}: Learning Persistent Voxel Feature Maps for Embodied Online Semantic Occupancy Mapping and Offline Refinement}

\author{
\textbf{
Ruoyu Wang$^{\dagger}$ \quad
Jiahan Li$^{\dagger}$ \quad
Yong Liu$^{*}$ \quad
Kexin Zhang$^{*}$
}\\[5pt]
\normalfont
APRIL Lab, Institute of Cyber-Systems and Control, Zhejiang University, China
}

\definecolor{ceiling}{RGB}{214, 122, 122}
\definecolor{floor}{RGB}{122, 214, 122}
\definecolor{wall}{RGB}{122, 122, 214}
\definecolor{window}{RGB}{214, 214, 122}
\definecolor{chair}{RGB}{214, 122, 214}
\definecolor{bed}{RGB}{161, 222, 245}
\definecolor{sofa}{RGB}{201, 173, 217}
\definecolor{table}{RGB}{245, 184, 122}
\definecolor{tvs}{RGB}{153, 184, 122}
\definecolor{furniture}{RGB}{122, 184, 184}
\definecolor{objects}{RGB}{82, 138, 199}

\iclrfinalcopy

\begin{document}
\maketitle

\begingroup
\renewcommand{\thefootnote}{}
\footnotetext{%
$^{\dagger}$ Equal contribution.
\quad
$^{*}$ Corresponding authors.
}
\endgroup

\fancyhead{}
\renewcommand{\headrulewidth}{0pt}

\input{sections/0_abstract}
\input{sections/1_introduction}
\input{sections/2_related_work}

\input{sections/3_methodology}
\input{sections/4_experiments}
\input{sections/5_conclusion}

\clearpage

\bibliography{ref}
\bibliographystyle{iclr2027_conference}

\clearpage
\appendix
\input{sections/7_appendix}

\end{document}

%% file: math_commands.tex
\usepackage{amsmath,amsfonts,bm}

\def\eqref#1{equation~\ref{#1}}

\def\1{\bm{1}}

\DeclareMathAlphabet{\mathsfit}{\encodingdefault}{\sfdefault}{m}{sl}
\SetMathAlphabet{\mathsfit}{bold}{\encodingdefault}{\sfdefault}{bx}{n}



%% file: sections/0_abstract.tex
\begin{abstract}
Semantic occupancy mapping supports embodied perception during exploration and
detailed scene understanding afterward, giving rise to two complementary
requirements: online incremental mapping for maintaining an up-to-date global
map under causal constraints, and offline global map refinement for improving
the completed scene representation with additional computation. We present
{\methodname}, a unified voxel-centric framework that connects both stages
through a persistent sparse voxel feature map. During exploration, the map
expands with newly observed regions and integrates aligned local features
through reliability-guided updates, while context-aware decoding produces the
current semantic occupancy map. After exploration, cached local features are
consolidated as complementary evidence and combined with the completed online
representation for multi-scale spatial refinement.
On EmbodiedOcc-ScanNet, {\methodname} achieves 67.12\% IoU and 58.40\% mIoU,
establishing new state-of-the-art performance for online incremental mapping.
With the local predictor fixed, our online mapping pipeline outperforms VEOcc
fusion by 3.70 IoU and 3.74 mIoU points, demonstrating substantial gains from
the proposed global mapping design. Offline global map refinement further
improves the completed map to 67.79\% IoU and 59.37\% mIoU using only the
observations acquired during exploration. Zero-shot deployment on a mobile
robot further demonstrates transferability to unseen environments, onboard
feasibility, and scalability to large-scale indoor mapping. Project page: \url{https://wryzju.github.io/VEOccpp/}.
\end{abstract}

%% file: sections/1_introduction.tex
\section{Introduction}

3D scene understanding~\citep{savva2019habitat,gu2024conceptgraphs,
rana2023sayplan,huang2023voxposer} is a fundamental capability for embodied
agents. 3D semantic occupancy maps provide an effective representation by
jointly modeling spatial occupancy and semantic categories in a fine-grained
volumetric space~\citep{mccormac2017semanticfusion,
yang2017semantic,song2017semantic,cao2022monoscene}. Practical embodied
applications place different demands on these maps during and after
exploration. During exploration, observations are sequentially integrated
under causal constraints to maintain an up-to-date global map for perception
and decision making. After exploration, the accumulated map and retained
evidence can be processed with additional computation to improve the final
scene representation. As shown in Fig.~\ref{fig:teaser}, we therefore
study \emph{online incremental mapping} and \emph{offline global map
refinement} as complementary stages of embodied semantic occupancy mapping.

Recent studies have extended semantic occupancy prediction from individual
observations to sequential mapping. Gaussian-centric approaches progressively
update scene-level representations~\citep{embodiedocc,embodiedoccpp,roboocc,
tgsformer,gpocc,gpoccpp}, whereas VEOcc~\citep{veocc} directly maintains an
expanding global voxel map. VEOcc performs long-term accumulation at the
semantic prediction level, without retaining the intermediate voxel features
for subsequent integration and spatial decoding. Beyond the choice of
accumulated representation, incorporating incoming observations and refining
a completed map involve different processing objectives. Sequential updates
prioritize timely integration, while refinement can devote dedicated
computation to broader spatial context across the accumulated scene. These
considerations motivate a persistent feature representation that supports both
online accumulation and subsequent global map refinement.

\input{figures/teaser}

To this end, we present {\methodname}, a unified voxel-centric framework that
connects both stages through a shared Persistent Voxel Feature Map. This sparse
map expands with exploration while retaining feature states independently of
their semantic readout. Reliability-Guided Incremental Update learns to
combine aligned local features with historical voxel states, while
Context-Aware Decoding incorporates neighboring voxel context to produce the
current occupancy map. After exploration, Local Evidence Aggregation
consolidates cached local features as complementary evidence. Global Spatial
Refinement then combines this evidence with the completed feature map and its
online semantic probabilities for multi-scale spatial reasoning. The resulting
refinement improves the final map without acquiring additional observations or
expanding the established map support.

Experiments on EmbodiedOcc-ScanNet demonstrate that the online stage alone
outperforms all compared methods, achieving 67.12\% IoU and 58.40\% mIoU.
With the local predictor held fixed, our online mapping pipeline improves IoU
and mIoU by 3.70 and 3.74 percentage points over VEOcc fusion, respectively,
demonstrating gains beyond improvements to the local predictor. Offline global
map refinement further raises the results to 67.79\% IoU and 59.37\% mIoU.
Beyond benchmark evaluation, zero-shot deployment on a mobile robot demonstrates
transferability to unseen real-world environments, onboard feasibility, and
scalability to large-scale indoor mapping.

Our main contributions are summarized as follows:

\begin{itemize}
    \item We develop a voxel-centric framework built upon a Persistent Voxel
    Feature Map that connects online incremental mapping with offline global
    map refinement. Reliability-guided updates and context-aware decoding
    enable feature-level scene accumulation and dynamic map expansion over time.

    \item We introduce offline global map refinement that combines cached local
    evidence with the completed online representation for multi-scale spatial
    reasoning, improving occupancy and semantic predictions without additional
    observations or changes to the established map support during refinement.

    \item Experiments demonstrate state-of-the-art online performance, and further improvements after
    refinement. Zero-shot robotic deployment demonstrates transferability,
    onboard feasibility, and scalability to large scale indoor environments.
\end{itemize}

%% file: figures/teaser.tex
\begin{figure}[t]
    \centering
    \captionsetup{font={small}}
    \includegraphics[width=\linewidth]{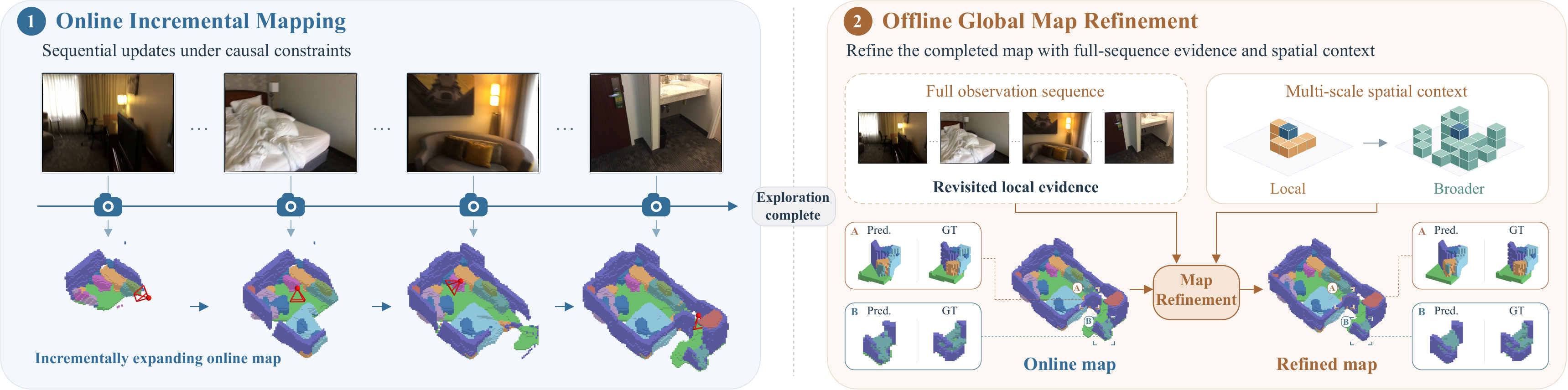}
\caption{
\textbf{Overview of the two-stage mapping setting in {\methodname}.}
Online incremental mapping sequentially integrates observations under causal
constraints, while offline global map refinement exploits complementary local
evidence and broader spatial context over the completed map.
Zoomed examples highlight representative improvements after refinement.
}
\label{fig:teaser}
\vspace{-10pt}
\end{figure}

%% file: sections/2_related_work.tex
\section{Related Work}

\subsection{Embodied Semantic Occupancy Mapping}
\label{subsec:related_embodied_occ}

Embodied semantic occupancy mapping extends frame-wise occupancy prediction to
sequential scene construction during exploration. EmbodiedOcc~\citep{embodiedocc}
introduces this setting with a Gaussian-centric global representation, followed
by methods that improve geometry modeling, temporal fusion, and representation
efficiency~\citep{embodiedoccpp,roboocc,tgsformer,gpocc,gpoccpp}. Recent works
further explore persistent Gaussian feature memory and long-horizon semantic
mapping~\citep{tgsformer,gemocc}, while FreeOcc~\citep{freeocc} studies
training-free open-vocabulary mapping. In contrast, VEOcc~\citep{veocc} adopts
a voxel-centric formulation with an incrementally expanding global map.
{\methodname} follows this direction but performs long-term accumulation in a
persistent voxel feature space and further connects online incremental mapping
with offline global map refinement.

\subsection{Voxel-Centric Persistent Scene Memory for Sequential 3D Perception}
\label{subsec:related_memory}

Voxel-centric persistent representations have been explored for sequential 3D
perception. NeuralRecon~\citep{neuralrecon} and
NeuralFusion~\citep{neuralfusion} maintain latent volumetric states for
incremental reconstruction and fusion, while
ST-Occ~\citep{leng2025occupancy}, StreamOcc~\citep{streamocc}, and
LMPOcc~\citep{lmpocc} extend related ideas to occupancy prediction through
spatiotemporal memory, recurrent voxel feature propagation, and global
occupancy priors. VEOcc~\citep{veocc} adopts an incrementally expanding voxel
map for embodied semantic occupancy, but performs long-term integration at the
prediction level. OccFiner~\citep{occfiner} focuses on offboard occupancy
refinement over completed prediction sequences, rather than causal online map
maintenance. In contrast, {\methodname} maintains a persistent sparse voxel
feature map as the shared scene representation for causal online mapping and
post-exploration refinement, preserving latent scene information beyond
semantic readout throughout both stages.

%% file: sections/3_methodology.tex
\section{Methodology}
\label{sec:methodology}

\subsection{Problem Formulation}
\label{subsec:problem_formulation}

Given a monocular image $\mathbf{I}_t \in \mathbb{R}^{H\times W\times 3}$
and geometric information $\mathbf{G}_t$, including camera intrinsics, pose,
and local voxel geometry, {\methodname} produces local and global semantic
occupancy predictions through
\begin{equation}
\label{eq:problem_formulation}
\begin{aligned}
\mathbf{Y}^{\mathrm{local}}_t
&=\mathcal{F}_{\mathrm{local}}(\mathbf{I}_t,\mathbf{G}_t),\\
(\mathcal{O}_t,\mathbf{Y}^{\mathrm{on}}_t)
&=\mathcal{F}_{\mathrm{online}}
(\mathcal{O}_{t-1},\mathbf{I}_t,\mathbf{G}_t),\\
\mathbf{Y}^{\mathrm{off}}_T
&=\mathcal{F}_{\mathrm{offline}}
\left(\mathcal{O}_T,\{(\mathbf{I}_t,\mathbf{G}_t)\}_{t=1}^{T}\right).
\end{aligned}
\end{equation}
Here, $\mathbf{Y}^{\mathrm{local}}_t$ is the frame-wise local prediction, and
$\mathcal{O}_t$ is the persistent global scene state from which the online map
$\mathbf{Y}^{\mathrm{on}}_t$ is decoded. Online incremental mapping processes
observations sequentially under causal constraints and expands the map support
as new regions are observed. After exploration ends at timestep $T$, offline
global map refinement uses the completed state and evidence retained from the
same sequence to produce $\mathbf{Y}^{\mathrm{off}}_T$ on the established
support.

\subsection{Method Overview}
\label{subsec:method_overview}

As illustrated in Fig.~\ref{fig:framework}, a Persistent Voxel Feature Map
connects the online and offline stages. The local predictor
(Sec.~\ref{subsec:local_prediction}) extracts voxel features from each frame.
Online incremental mapping (Sec.~\ref{subsec:online_mapping}) aligns and
integrates these features into the persistent map, from which semantic
occupancy is decoded. Offline global map refinement
(Sec.~\ref{subsec:offline_refinement}) subsequently combines the completed
online representation with cached local evidence for multi-scale spatial
refinement.

\subsection{Local Occupancy Prediction}
\label{subsec:local_prediction}

We adopt the local predictor of VEOcc~\citep{veocc} without architectural
modifications. Given $\mathbf{I}_t$ and $\mathbf{G}_t$, it extracts image
features and incorporates a monocular depth prior~\citep{depthanythingv2} to
produce a local voxel feature volume $\mathbf{F}_t$ and the local occupancy
prediction $\mathbf{Y}^{\mathrm{local}}_t$. The features before the
occupancy head are retained for online accumulation and offline evidence
aggregation.
\input{figures/method}
\subsection{Online Incremental Mapping}
\label{subsec:online_mapping}

\paragraph{Persistent Voxel Feature Map.}
The global scene state $\mathcal{O}_t$ comprises a persistent sparse voxel
feature map
$\mathcal{M}_t=\{(k,\mathbf{h}_t(k))\mid k\in\mathcal{K}_t\}$
and auxiliary voxel-wise reliability statistics. Here, $\mathcal{K}_t$
contains the stored voxel coordinates in the world frame, and
$\mathbf{h}_t(k)\in\mathbb{R}^{C}$ is the feature associated with voxel $k$.
The map retains accumulated scene features independently of their semantic
readout, allowing the same representation to support subsequent online updates
and offline refinement. Its support grows with exploration without requiring
a predefined global scene boundary.

\paragraph{Observation-to-Map Alignment.}
For each incoming frame, we select geometrically visible local voxels and
apply a lightweight pose-conditioned adaptation to their features. Let
$\widehat{\mathbf{F}}_t$ denote the adapted local features,
$\mathcal{S}_t$ the selected voxel set, and $\mathbf{w}_{t,i}$ the world
coordinate of voxel $i\in\mathcal{S}_t$. With global voxel size $\delta$,
quantization gives the active map coordinates
$\mathcal{A}_t=\{\lfloor\mathbf{w}_{t,i}/\delta\rfloor\mid
i\in\mathcal{S}_t\}$.

To account for misalignment between the local and global voxel lattices, we
sample features at each active global voxel center
$\mathbf{q}_k=\delta(k+\frac{1}{2})$:
\begin{equation}
\label{eq:voxel_center_sampling}
\mathbf{x}_t(k)
=
\operatorname{TriSample}
\left(
\widehat{\mathbf{F}}_t,\mathbf{q}_k;\mathbf{G}_t
\right),
\qquad
k\in\mathcal{A}_t .
\end{equation}
The sampling operator uses $\mathbf{G}_t$ to express the world-coordinate
query in the local voxel lattice before trilinear interpolation.
The resulting $\mathbf{x}_t(k)$ is the aligned observation feature.
Pose-conditioned adaptation is detailed in
Appendix~\ref{app:alignment_details}.

\paragraph{Reliability-Guided Incremental Update.}
Given $\mathcal{A}_t$, we expand the map support and update its feature
states as
\begin{equation}
\label{eq:memory_state_transition}
\mathcal{K}_t=\mathcal{K}_{t-1}\cup\mathcal{A}_t,
\qquad
\mathbf{h}_t(k)=
\begin{cases}
\mathbf{x}_t(k),
& k\in\mathcal{A}_t\setminus\mathcal{K}_{t-1},\\
\mathcal{U}\!\left(\mathbf{h}_{t-1}(k),\mathbf{x}_t(k)\right),
& k\in\mathcal{A}_t\cap\mathcal{K}_{t-1},\\
\mathbf{h}_{t-1}(k),
& k\in\mathcal{K}_{t-1}\setminus\mathcal{A}_t.
\end{cases}
\end{equation}
New voxels are initialized from the aligned features, while voxels outside the
current observation retain their previous states. For re-observed voxels,
$\mathcal{U}$ adapts the incoming feature and determines its contribution to
the persistent state.
Following VEOcc~\citep{veocc}, we estimate the current reliability $r_t(k)$
from depth and image-boundary cues and maintain the historical average
reliability $\bar r_{t-1}(k)$. Together with semantic uncertainty $E_t(k)$
from the local prediction and relations between the historical and current
features, these quantities form an update descriptor
$\boldsymbol{\xi}_t(k)$. Its complete construction is provided in
Appendix~\ref{app:memory_details}.

The descriptor conditions a corrected candidate feature and a channel-wise
update gate:
\begin{equation}
\label{eq:update_gate}
\begin{aligned}
\widetilde{\mathbf{x}}_t(k)
&=
\mathbf{x}_t(k)
+
\Delta_c\!\left(\boldsymbol{\xi}_t(k)\right),\\
\mathbf{g}_t(k)
&=
\sigma\!\left(
\operatorname{logit}\!\left(
\frac{r_t(k)}
{r_t(k)+\bar r_{t-1}(k)+\epsilon}
\right)
+
\Delta_g\!\left(\boldsymbol{\xi}_t(k)\right)
\right).
\end{aligned}
\end{equation}
Here, $\Delta_c$ and $\Delta_g$ are learned residual functions.
The candidate branch adjusts the incoming feature, while the gate balances
its contribution against the historical state. The reliability ratio anchors
this balance to observation quality, and $\Delta_g$ learns
descriptor-dependent corrections. The resulting update is
\begin{equation}
\label{eq:persistent_update}
\mathcal{U}\!\left(
\mathbf{h}_{t-1}(k),\mathbf{x}_t(k)
\right)
=
\left(1-\mathbf{g}_t(k)\right)\odot\mathbf{h}_{t-1}(k)
+
\mathbf{g}_t(k)\odot\widetilde{\mathbf{x}}_t(k).
\end{equation}

\paragraph{Context-Aware Decoding.}
Feature updates are voxel-wise, whereas occupancy prediction also benefits
from interactions among neighboring map voxels. A lightweight sparse
convolutional head $\mathcal{H}_{\mathrm{map}}$ therefore decodes persistent
features over the entire current support $\mathcal{K}_t$ into semantic logits
$\mathbf{L}^{\mathrm{on}}_t$, yielding semantic probabilities
$\mathbf{P}^{\mathrm{on}}_t$ and the online occupancy map
$\mathbf{Y}^{\mathrm{on}}_t$, with the former retained for subsequent offline
refinement.

\subsection{Offline Global Map Refinement}
\label{subsec:offline_refinement}

After exploration, we refine the completed online representation using cached evidence from the
same image sequence. The spatial support $\mathcal{K}_T$ remains fixed, while occupancy and semantic
predictions are revised through multi-scale spatial processing. This offline refinement consists of
Local Evidence Aggregation and Global Spatial Refinement, described below.

\paragraph{Local Evidence Aggregation.}
Local voxel features and their geometry are cached during online
processing. After exploration, we retrieve these features and align them to
$\mathcal{K}_T$ without rerunning the local predictor. For each mapped voxel
$k$, the aligned features form
$\mathcal{V}_k=\{\mathbf{o}_{k,j}\}_{j=1}^{N_k}$, where
$\mathbf{o}_{k,j}\in\mathbb{R}^{C}$ and $N_k$ is the number of valid features.
Each feature undergoes a residual transformation and receives a scalar weight
normalized over the observations associated with the same voxel:
\begin{equation}
\label{eq:offline_aggregation}
\mathbf{v}_{k,j}
=
\mathbf{o}_{k,j}+\psi(\mathbf{o}_{k,j}),
\qquad
w_{k,j}
=
\operatorname{Softmax}_{j}\!\left(\rho(\mathbf{v}_{k,j})\right),
\qquad
\mathbf{u}_k
=
\sum_{j=1}^{N_k}w_{k,j}\mathbf{v}_{k,j}.
\end{equation}
Here, $\psi$ and $\rho$ are lightweight MLPs. The aggregated feature
$\mathbf{u}_k$ supplies complementary local evidence for global refinement.

\paragraph{Global Spatial Refinement.}
For each mapped voxel, we concatenate the persistent feature, online semantic
probability, and aggregated local evidence as
$\mathbf{z}_k=
[\mathbf{h}_T(k),\mathbf{p}^{\mathrm{on}}_T(k),\mathbf{u}_k]$,
where
$\mathbf{p}^{\mathrm{on}}_T(k)=
\operatorname{Softmax}(\mathbf{L}^{\mathrm{on}}_T(k))$.
These inputs provide the accumulated scene features, their current semantic
interpretation, and complementary frame-wise evidence, respectively.
A sparse U-Net combines them across spatial scales to predict
\begin{equation}
\label{eq:offline_refinement}
\mathbf{L}^{\mathrm{off}}_T
=
\mathcal{R}_{\mathrm{ref}}
\left(
\{\mathbf{z}_k\}_{k\in\mathcal{K}_T}
\right).
\end{equation}
Compared with the lightweight online decoder, this multi-scale processing
incorporates broader spatial context into the prediction of each mapped voxel.
The refined features are directly decoded into final semantic logits
$\mathbf{L}^{\mathrm{off}}_T$. Semantic classification then produces
$\mathbf{Y}^{\mathrm{off}}_T$, updating occupancy and category labels while
preserving the spatial support.

%% file: figures/method.tex
\begin{figure}[t]
    \centering
    \captionsetup{font={small}}
    \includegraphics[width=\linewidth]{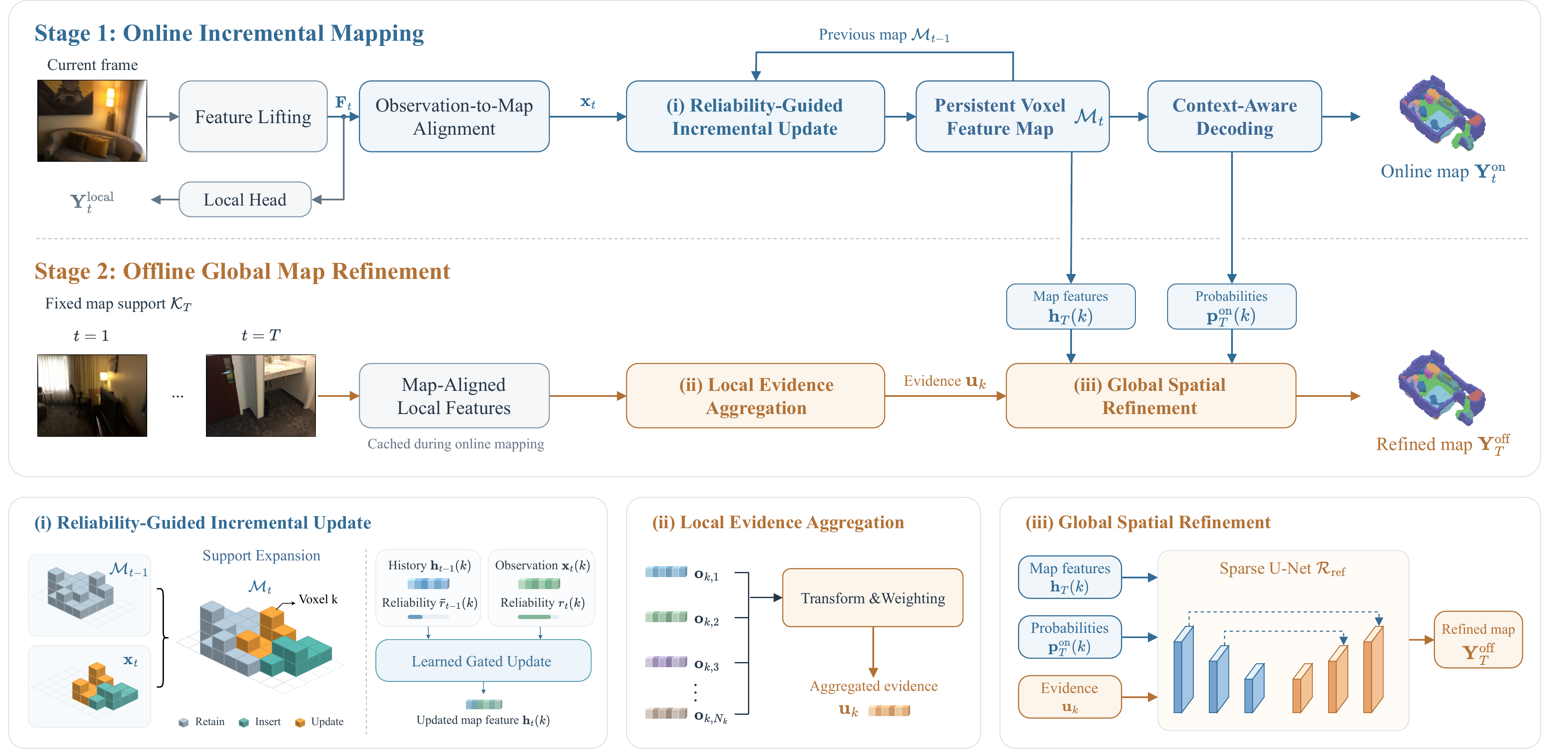}
\caption{
\textbf{Overall framework of {\methodname}.}
During online incremental mapping, local voxel features are aligned with the
global map and accumulated in a Persistent Voxel Feature Map through
reliability-guided updates, while Context-Aware Decoding produces the current
occupancy map. During offline global map refinement, Local Evidence Aggregation
consolidates cached local features as complementary evidence, which is combined
with the completed online representation for Global Spatial Refinement.
}
\label{fig:framework}
\vspace{-10pt}
\end{figure}

%% file: sections/4_experiments.tex
\section{Experiments}
\label{sec:experiments}

\input{tables/local_pred}
\input{tables/global_pred}

\subsection{Experimental Setup}
\label{subsec:experimental_setup}

\paragraph{Datasets and Metrics.}
We evaluate {\methodname} on two benchmarks: Occ-ScanNet~\citep{occscannet}
for local occupancy prediction and EmbodiedOcc-ScanNet~\citep{embodiedocc}
for online incremental mapping and offline global map refinement.
Occ-ScanNet provides voxelized annotations in a
$60\times60\times36$ grid covering a
$4.8\,\mathrm{m}\times4.8\,\mathrm{m}\times2.88\,\mathrm{m}$ local volume,
while EmbodiedOcc-ScanNet extends the annotations to complete scenes for
sequential global mapping evaluation. We report occupancy IoU and semantic mIoU, together with
runtime and memory usage for efficiency evaluation.

\paragraph{Implementation Details.}
{\methodname} is trained in three stages. We first train the VEOcc local
occupancy predictor with a larger learning rate while keeping its architecture
unchanged. We then freeze the local predictor and train the online incremental
mapping modules. Finally, both the local predictor and online mapper are fixed,
and only the refinement modules are optimized on completed scene maps. Training
is conducted on four NVIDIA RTX 3090 GPUs and takes approximately 28 hours in
total. Additional training and implementation details are provided in
Appendix~\ref{app:training_details}.

\subsection{Results on Public Benchmarks}
\label{subsec:sota}

\paragraph{Local Occupancy Prediction.}
We first evaluate frame-wise occupancy prediction on Occ-ScanNet. Increasing
the learning rate improves the unchanged local predictor to 66.98\% IoU and
58.17\% mIoU, achieving the best performance among the compared embodied
occupancy mapping methods in Table~\ref{tab:local_pred}. 
Further analysis of the learning rate is provided in
Appendix~\ref{app:local_lr}. 
The resulting local
features provide a strong basis for subsequent online mapping and offline refinement.

\paragraph{Global Occupancy Mapping.}
\textbf{(i) Online Mapping Performance.}
As shown in Table~\ref{tab:global_pred}, {\methodname} establishes new
state-of-the-art performance on EmbodiedOcc-ScanNet, achieving 67.12\% IoU
and 58.40\% mIoU in the online setting. It surpasses GPOcc++ by 4.62 IoU and
0.99 mIoU points, despite GPOcc++ adopting a substantially larger VGGT
backbone with large-scale pretraining. The strong performance obtained before
any offline refinement highlights the effectiveness of our online incremental
mapping design.
\textbf{(ii) Mapping Gains with a Fixed Local Predictor.}
To isolate the contribution of global mapping, we additionally report
VEOcc$^\dagger$, which uses the same local predictor as {\methodname} while
retaining the analytic prediction-level fusion of VEOcc. This stronger local
predictor improves VEOcc from 62.21/53.00 to 63.42/54.66 IoU/mIoU.
With the local predictor fixed, {\methodname} further reaches 67.12/58.40,
yielding gains of 3.70 IoU and 3.74 mIoU points. Improvements are observed
across all 11 semantic categories, demonstrating a substantial advantage of
the proposed online mapping pipeline over the VEOcc formulation.
\textbf{(iii) Offline Refinement.}
Starting from the completed online map, offline global map refinement further
improves performance to 67.79\% IoU and 59.37\% mIoU. Under the same
observations and fixed spatial support, our offline refinement yields gains of
0.67 IoU and 0.97 mIoU points over the online result, with improvements across
nearly all semantic categories. These consistent gains demonstrate that the
proposed refinement effectively exploits previously acquired evidence to enhance
global map quality without additional observations or spatial expansion.

\input{figures/embodied_vis_self}
\input{figures/embodied_vis_compare}

\paragraph{Qualitative Results.}
Fig.~\ref{fig:online_offline_vis} illustrates progressive map construction
during exploration and subsequent offline refinement. The online map expands
as observations arrive, while offline global map refinement reduces residual
geometric and semantic artifacts in the highlighted regions.
Fig.~\ref{fig:online_compare} further compares online incremental mapping
with GPOcc and VEOcc on representative EmbodiedOcc-ScanNet scenes.
{\methodname} produces more complete object structures and cleaner semantic
boundaries, complementing the quantitative global mapping results.

\input{figures/realworld-vis}
\input{tables/efficiency}
\paragraph{Efficiency Analysis.}
We evaluate {\methodname} in full precision on a single RTX 3090. As shown
in Table~\ref{tab:efficiency}, the online incremental mapping pipeline averages
152.44\,ms/frame, corresponding to approximately 6.6 FPS, while offline
global map refinement averages 319.03\,ms per 30-frame scene. This refinement
cost is only about 7.0\% of the cumulative online processing time for the same
scene. The persistent map requires 424 bytes per voxel, with moderate GPU memory
peaks in both stages, providing a practical full-precision deployment baseline.
We further evaluate a mixed-FP16 deployment configuration. As shown in
Table~\ref{tab:efficiency}, this configuration reduces the overall deployment
cost with negligible accuracy degradation and is therefore adopted for the
subsequent real-world experiments. The slightly higher offline GPU memory peak
mainly reflects the additional integer indexing overhead of the Native
sparse-convolution backend, which outweighs the memory savings from
half-precision features.

\subsection{Zero-Shot Real-World Robotic Experiments}
\label{subsec:realworld}

\paragraph{Robotic Platform.}
We evaluate zero-shot transfer and onboard feasibility on an AgileX UMR
mobile robot equipped with an Intel RealSense D435i, a Livox MID-360 LiDAR
with IMU for localization, and an NVIDIA Jetson Orin 16GB for computation
(Fig.~\ref{fig:robot_platform}). 
\input{figures/robot_platform}
Occupancy prediction uses monocular RGB
images, while camera poses are provided by the LiDAR--inertial localization
system. The benchmark-trained model is deployed without scene-specific
fine-tuning or adaptation. All real-world experiments use mixed FP16 together
with TensorRT-accelerated monocular depth estimation.

\paragraph{Real-World Mapping Scenarios.}
We consider two unseen indoor environments to examine deployment across
different scene scales: a furnished Reading Room and a complete Laboratory
Floor. As summarized in Table~\ref{tab:realworld}, their trajectories span
31.31\,m and 235.15\,m, respectively, covering compact room-scale exploration
and extended floor-scale mapping. Motion-gated keyframe selection with a
minimum interval of one second yields 91 and 351 processed frames,
respectively.

\paragraph{Qualitative Results.}
Fig.~\ref{fig:realworld_vis} shows the maps produced without scene-specific
adaptation. In the Reading Room, online incremental mapping captures the room
layout and recognizable furniture structures. Offline global map refinement
further reduces surface fragmentation in the highlighted object and wall
regions while retaining the overall layout. On the Laboratory Floor, the
online map represents connected corridor and room regions over the longer
trajectory, with identifiable structural and object-level semantics in the
enlarged views. Additional real-world comparisons with VEOcc across multiple
mapping sequences are provided in
Appendix~\ref{app:realworld_online_comparison}.

\paragraph{Deployment Efficiency.}
We measure runtime and memory during sequential replay of the recorded
sequences on the onboard device. Table~\ref{tab:realworld} evaluates the
complete pipeline in both room-scale and floor-scale environments. In both
settings, online mapping takes less than one second per keyframe on average,
including local-feature cache writes, with average latencies of 0.630 and
0.764\,s/frame for the Reading Room and Laboratory Floor, respectively.
The corresponding persistent maps occupy 35.07 and 215.69\,MiB, while
complete offline refinement takes 4.29 and 25.07\,s. In both environments,
the offline refinement cost remains below 10\% of the cumulative online
processing time. Even for the full-floor sequence, peak process RSS remains
5.136\,GiB during online mapping and 7.806\,GiB during refinement. These
results show that {\methodname} supports both compact room-scale and
long-horizon floor-scale mapping on the same embedded platform, demonstrating
its practical efficiency and scalability for onboard embodied scene
understanding. A finer-grained analysis of latency, process memory, and persistent-map growth
throughout the Laboratory Floor sequence is provided in
Appendix~\ref{app:deployment_scaling}.

\subsection{Ablation Studies}
\label{subsec:ablations}

\input{tables/deployment}
\input{tables/ablation_embodied}

\paragraph{Online Incremental Mapping.}
\label{subsec:online_ablation}

Table~\ref{tab:online_ablation} analyzes the main components of online
incremental mapping under the same local predictor. Removing persistent memory
causes a substantial drop to 61.33/52.03 IoU/mIoU, while replacing analytic
prediction-level fusion with a Persistent Voxel Feature Map and cumulative
mean update improves performance from 63.42/54.66 to 66.25/57.49. Building
on this representation, reliability-guided gating and candidate adaptation
provide complementary improvements when combined in the full
Reliability-Guided Incremental Update, reaching 67.12/58.40. Replacing
Context-Aware Decoding with pointwise prediction reduces performance to
66.51/57.81, further showing the benefit of spatial context when decoding the feature map.

\paragraph{Offline Global Map Refinement.}
\label{subsec:offline_ablation}

Table~\ref{tab:offline_ablation} examines the contributions of the three input
sources and spatial refinement. Applying Global Spatial Refinement to online
semantic probabilities alone improves the completed map from 67.12/58.40 to
67.36/58.62 IoU/mIoU. Adding either persistent map features or learned local
evidence further raises performance to 67.54/59.03 and 67.56/58.94,
respectively, showing that both provide useful information beyond the online
semantic prediction. Combining persistent features with mean-aggregated local
evidence reaches 67.59/59.05, while learned Local Evidence Aggregation yields
the best result of 67.79/59.37. Replacing Global Spatial Refinement with
pointwise prediction reduces performance to 67.21/58.68. Overall, persistent
map features and local evidence provide complementary cues, and spatial
refinement is important for effectively exploiting them over the completed map.

%% file: tables/local_pred.tex
\begin{table*}[t]
    \captionsetup{font={small}}
    \setlength{\tabcolsep}{0.0035\linewidth}
    \centering
    \caption{\textbf{Local Occupancy Prediction} results of embodied occupancy mapping methods
on Occ-ScanNet.
    The best results are highlighted in \textbf{bold}, and the second-best
    results are \underline{underlined}.}
    \label{tab:local_pred}

    \resizebox{\linewidth}{!}{
    \begin{tabular}{l |c c |c c c c c c c c c c c}
        \toprule
        Method
        & \makecell[c]{IoU}
        & \makecell[c]{mIoU}
        & \rotatebox{90}{\textcolor{ceiling}{$\blacksquare$} ceiling}
        & \rotatebox{90}{\textcolor{floor}{$\blacksquare$} floor}
        & \rotatebox{90}{\textcolor{wall}{$\blacksquare$} wall}
        & \rotatebox{90}{\textcolor{window}{$\blacksquare$} window}
        & \rotatebox{90}{\textcolor{chair}{$\blacksquare$} chair}
        & \rotatebox{90}{\textcolor{bed}{$\blacksquare$} bed}
        & \rotatebox{90}{\textcolor{sofa}{$\blacksquare$} sofa}
        & \rotatebox{90}{\textcolor{table}{$\blacksquare$} table}
        & \rotatebox{90}{\textcolor{tvs}{$\blacksquare$} tvs}
        & \rotatebox{90}{\textcolor{furniture}{$\blacksquare$} furniture}
        & \rotatebox{90}{\textcolor{objects}{$\blacksquare$} objects} \\
        \midrule

        EmbodiedOcc~\citep{embodiedocc}
        & 53.95 & 45.48 & 40.90 & 50.80 & 41.90 & 33.00
        & 41.20 & 55.20 & 61.90 & 43.80 & 35.40 & 53.50 & 42.90 \\

        EmbodiedOcc++~\citep{embodiedoccpp}
        & 54.90 & 46.20 & 36.40 & 53.10 & 41.80 & 34.40
        & 42.90 & 57.30 & 64.10 & 45.20 & 34.80 & 54.20 & 44.10 \\

        RoboOcc~\citep{roboocc}
        & 56.48 & 47.67 & 45.36 & 53.49 & 44.35 & 34.81
        & 43.38 & 56.93 & 63.35 & 46.35 & 36.12 & 55.48 & 44.78 \\

        TGSFormer~\citep{tgsformer}
        & 64.42 & 54.73 & 54.20 & 61.30 & 50.50 & 41.30
        & 49.50 & 64.50 & 69.30 & 52.40 & 47.10 & 63.00 & 48.80 \\

        GPOcc (VGGT)~\citep{gpocc}
        & 63.14 & 56.19 & 51.67 & 59.93 & \textbf{52.07}
        & \underline{46.44} & 51.35 & 64.45 & 69.47 & 54.30
        & \textbf{51.76} & 63.29 & \underline{53.36} \\

        GPOcc++ (VGGT)~\citep{gpoccpp}
        & 63.72 & \underline{57.41} & \textbf{56.62} & 61.40 & 50.63
        & \textbf{48.13} & \textbf{53.12} & \underline{66.34}
        & \underline{71.05} & \textbf{56.48} & 48.95
        & \underline{64.48} & \textbf{54.22} \\

        VEOcc~\citep{veocc}
        & \underline{64.55} & 55.49 & 53.20 & \underline{64.00}
        & 49.00 & 42.90 & 50.80 & 65.50 & 70.50 & 53.70
        & 48.30 & 62.60 & 50.00 \\

        \midrule

        {\methodname}(Ours)
        & \textbf{66.98} & \textbf{58.17} & \underline{55.70}
        & \textbf{67.30} & \underline{51.60} & 46.00
        & \underline{53.00} & \textbf{67.40} & \textbf{72.70}
        & \underline{56.10} & \underline{51.20}
        & \textbf{65.40} & 53.30 \\

        \bottomrule
    \end{tabular}
    }
\end{table*}

%% file: tables/global_pred.tex
\begin{table*}[t]
    \captionsetup{font={small}}
    \setlength{\tabcolsep}{0.0035\linewidth}
    \centering
    \caption{\textbf{Global Occupancy Mapping} results on EmbodiedOcc-ScanNet.
    The best results are highlighted in \textbf{bold}, and the second-best
    results are \underline{underlined}. $\dagger$ denotes VEOcc using the same
    local predictor as {\methodname}.}
    \label{tab:global_pred}

    \resizebox{\linewidth}{!}{
    \begin{tabular}{l |c c |c c c c c c c c c c c}
        \toprule
        Method
        & \makecell[c]{IoU}
        & \makecell[c]{mIoU}
        & \rotatebox{90}{\textcolor{ceiling}{$\blacksquare$} ceiling}
        & \rotatebox{90}{\textcolor{floor}{$\blacksquare$} floor}
        & \rotatebox{90}{\textcolor{wall}{$\blacksquare$} wall}
        & \rotatebox{90}{\textcolor{window}{$\blacksquare$} window}
        & \rotatebox{90}{\textcolor{chair}{$\blacksquare$} chair}
        & \rotatebox{90}{\textcolor{bed}{$\blacksquare$} bed}
        & \rotatebox{90}{\textcolor{sofa}{$\blacksquare$} sofa}
        & \rotatebox{90}{\textcolor{table}{$\blacksquare$} table}
        & \rotatebox{90}{\textcolor{tvs}{$\blacksquare$} tvs}
        & \rotatebox{90}{\textcolor{furniture}{$\blacksquare$} furniture}
        & \rotatebox{90}{\textcolor{objects}{$\blacksquare$} objects} \\
        \midrule

        EmbodiedOcc~\citep{embodiedocc}
        & 51.52 & 42.53
        & 22.70 & 44.60 & 37.40 & 38.00
        & 50.10 & 56.70 & 59.70 & 35.40
        & 38.40 & 52.00 & 32.90 \\

        EmbodiedOcc++~\citep{embodiedoccpp}
        & 52.20 & 43.60
        & 27.90 & 43.90 & 38.70 & 40.60
        & 49.00 & 57.90 & 59.20 & 36.80
        & 37.80 & 53.50 & 34.10 \\

        RoboOcc~\citep{roboocc}
        & 53.30 & 44.05
        & 21.94 & 44.57 & 39.54 & 38.48
        & 51.28 & 57.04 & 63.09 & 36.70
        & 43.05 & 54.42 & 34.38 \\

        TGSFormer~\citep{tgsformer}
        & 54.42 & 45.29
        & 31.00 & 39.90 & 39.90 & 32.70
        & 52.00 & 66.10 & 53.00 & 47.30
        & 38.70 & 59.60 & 37.90 \\

        GPOcc (VGGT)~\citep{gpocc}
        & 61.41 & 55.39
        & 42.61 & 51.35 & 51.49 & 48.72
        & 54.32 & 67.91 & 70.73 & 52.94
        & \underline{54.75} & 64.76 & 49.67 \\

        GPOcc++ (VGGT)~\citep{gpoccpp}
        & 62.50 & 57.41
        & \textbf{48.00} & 52.14 & \textbf{54.70} & \textbf{53.39}
        & 54.71 & 68.82 & 71.03 & 54.27
        & \textbf{55.23} & 66.81 & 52.42 \\

        VEOcc~\citep{veocc}
        & 62.21 & 53.00
        & 29.90 & 62.60 & 45.60 & 44.60
        & 54.10 & 68.10 & 69.70 & 54.30
        & 45.90 & 61.60 & 46.40 \\

        VEOcc$^\dagger$~\citep{veocc}
        & 63.42 & 54.66
        & 33.20 & 63.40 & 47.00 & 46.60
        & 55.40 & 69.00 & 71.60 & 55.90
        & 47.10 & 63.40 & 48.60 \\

        \midrule

        {\methodname} (Ours, Online)
        & \underline{67.12} & \underline{58.40}
        & \underline{43.09} & \underline{66.87} & 51.95 & 49.11
        & \underline{57.87} & \underline{70.49} & \underline{73.99}
        & \underline{60.01} & 48.44
        & \underline{67.28} & \underline{53.34} \\

        {\methodname} (Ours, Offline)
        & \textbf{67.79} & \textbf{59.37}
        & 42.77 & \textbf{68.76} & \underline{52.12} & \underline{49.51}
        & \textbf{58.68} & \textbf{70.74} & \textbf{74.65}
        & \textbf{61.48} & 52.37
        & \textbf{68.01} & \textbf{54.01} \\

        \bottomrule
    \end{tabular}
    }
\vspace{-10pt}
\end{table*}

%% file: figures/embodied_vis_self.tex
\begin{figure}[t]
    \centering
    \captionsetup{font={small}}
    \includegraphics[width=\linewidth]{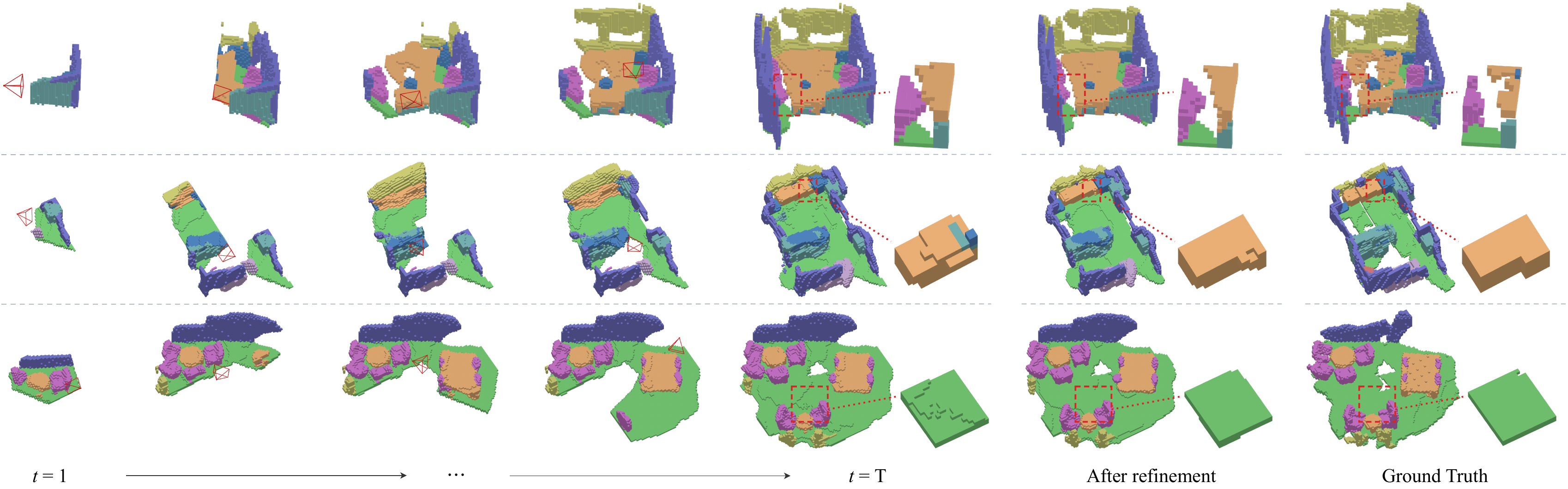}
    \caption{
    \textbf{Progressive online incremental mapping and offline map refinement.}
    Each row shows the map evolution from $t=1$ to $t=T$, followed by the
    refined result and ground truth. Red dashed boxes indicate the enlarged
    regions.
    }
    \label{fig:online_offline_vis}
\end{figure}

%% file: figures/embodied_vis_compare.tex
\begin{figure}[t]
    \centering
    \captionsetup{font={small}}
    \includegraphics[width=\linewidth]{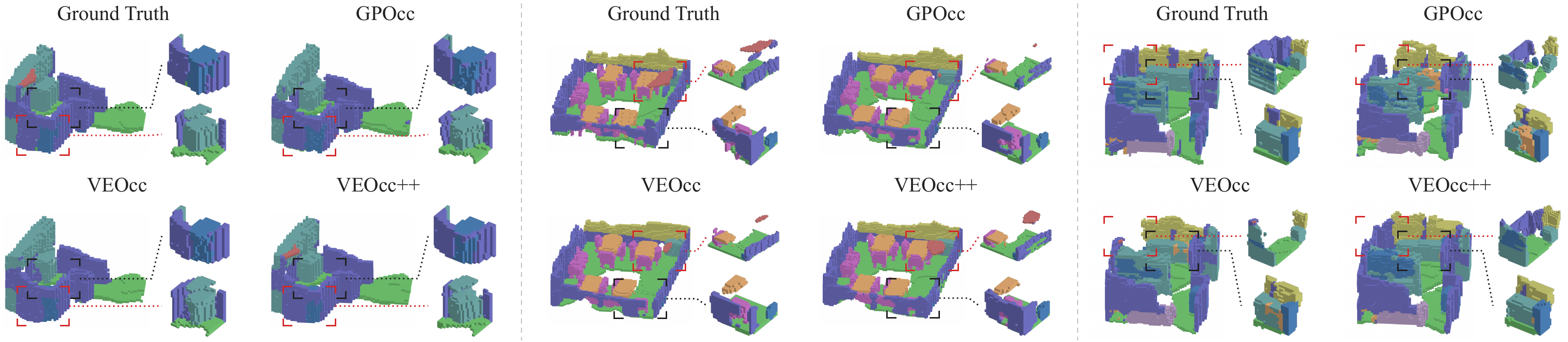}
    \caption{
    \textbf{Qualitative comparison of online incremental mapping.}
    Highlighted regions are enlarged for detailed comparison.
    }
    \label{fig:online_compare}
    \vspace{-10pt}
\end{figure}

%% file: figures/realworld-vis.tex
\begin{figure}[t]
    \centering
    \captionsetup{font={small}}
    \includegraphics[width=\linewidth]{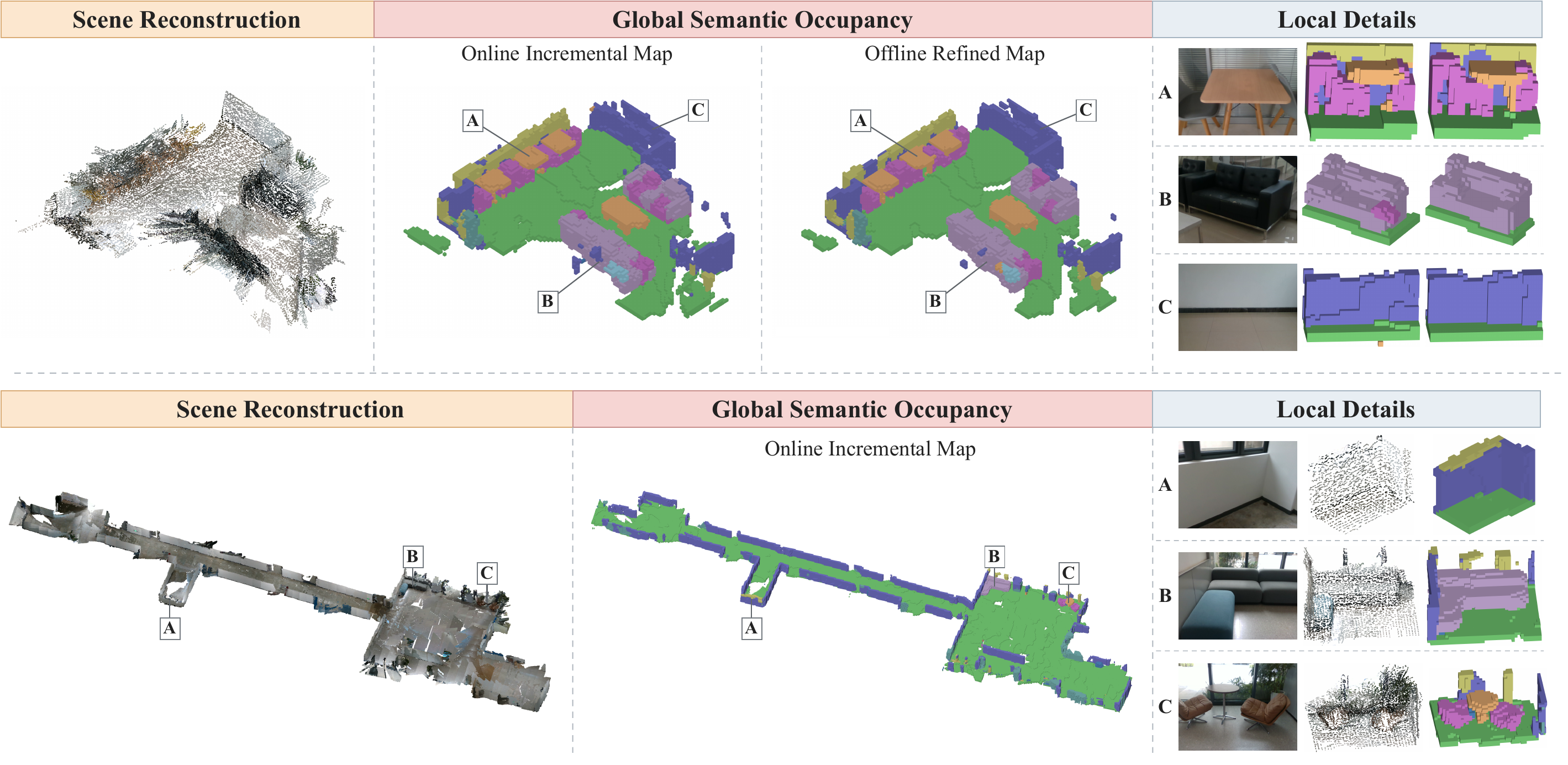}
    \caption{
    \textbf{Zero-shot real-world mapping results.}
    The upper and lower rows show the Reading Room and Laboratory Floor,
    respectively. Scene reconstructions are provided only for spatial reference
    and are not used by the occupancy predictor. The global maps demonstrate
    online incremental mapping at different scene scales, while offline
    refinement further improves the Reading Room result. A--C highlight
    representative local regions.
    }
    \label{fig:realworld_vis}
    \vspace{-10pt}
\end{figure}

%% file: tables/efficiency.tex
\begin{table}[H]
\centering
\captionsetup{font=small}
\caption{\textbf{Efficiency and precision comparison} on EmbodiedOcc-ScanNet.
Runtime and GPU memory are measured on a single RTX 3090.}
\label{tab:efficiency}

\scriptsize
\setlength{\tabcolsep}{2.2pt}
\begin{tabular}{l|ccc|cc|c|cccc}
\toprule
Setting
& \makecell{Online\\s/frame}
& \makecell{Online\\s/scene}
& \makecell{Offline\\s/scene}
& \makecell{Online GPU\\GiB}
& \makecell{Offline GPU\\GiB}
& \makecell{Map\\B/voxel}
& \makecell{Online\\IoU}
& \makecell{Online\\mIoU}
& \makecell{Offline\\IoU}
& \makecell{Offline\\mIoU} \\
\midrule

FP32
& 0.152
& 4.57
& 0.319
& 1.753
& 3.169
& 424
& 67.12
& 58.40
& 67.79
& 59.37 \\

Mixed FP16
& 0.120
& 3.60
& 0.224
& 1.558
& 3.252
& 232
& 67.08
& 58.36
& 67.77
& 59.35 \\

\bottomrule
\end{tabular}
\vspace{-10pt}
\end{table}

%% file: figures/robot_platform.tex
\begin{wrapfigure}{r}{0.30\linewidth}
    \vspace{-8pt}
    \centering
    \captionsetup{font={scriptsize}}
    \includegraphics[width=\linewidth]{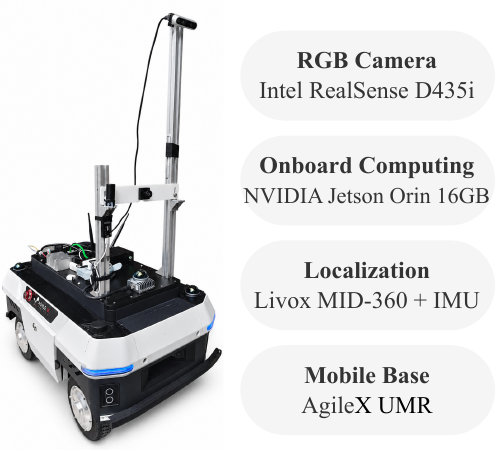}
    \caption{
    Real-world robotic platform.
    }
    \label{fig:robot_platform}
    \vspace{-25pt}
\end{wrapfigure}

%% file: tables/deployment.tex
\begin{table}[t]
\centering
\captionsetup{font=small}
\caption{\textbf{Real-world deployment results} in two indoor environments.
Map extent denotes the horizontal bounding box of the final stored voxel map.
RSS reports the peak process memory usage in each stage, while Local Cache
denotes the cached local features stored for offline refinement.}
\label{tab:realworld}

\scriptsize
\setlength{\tabcolsep}{1.3pt}
\begin{tabular}{l|ccc|ccc|cccc}
\toprule
Scene
& \makecell{Travel\\Dist. (m)}
& \makecell{Map Extent\\(m)}
& \makecell{Key\\Frames}
& \makecell{Online Avg.\\(s/frame)}
& \makecell{Online Time\\(s/scene)}
& \makecell{Offline\\Refine (s)}
& \makecell{Online RSS\\(GiB)}
& \makecell{Offline RSS\\(GiB)}
& \makecell{Feature Map\\(MiB)}
& \makecell{Local Cache\\(MiB)} \\
\midrule

Reading Room
& 31.31
& $10.0\times9.0$
& 91
& 0.630
& 57.32
& 4.29
& 4.172
& 4.340
& 35.07
& 151.55 \\

Laboratory Floor
& 235.15
& $57.3\times15.6$
& 351
& 0.764
& 268.08
& 25.07
& 5.136
& 7.806
& 215.69
& 576.50 \\

\bottomrule
\end{tabular}
\end{table}

%% file: tables/ablation_embodied.tex
\begin{table}[t]
\centering
\captionsetup{font=small}

\begin{minipage}[t]{0.495\linewidth}
\vspace{0pt}
\centering
\scriptsize

\caption{\textbf{Online mapping ablation.}
PFM: persistent feature map; CA: candidate adaptation;
RG: reliability-guided gating; CAD: context-aware decoding.}
\label{tab:online_ablation}

\setlength{\tabcolsep}{2.0pt}

\begin{tabular}{
    @{}
    >{\raggedright\arraybackslash}p{0.29\linewidth}|
    C{0.065\linewidth}
    C{0.065\linewidth}
    C{0.065\linewidth}
    C{0.065\linewidth}|
    C{0.10\linewidth}
    C{0.11\linewidth}
    @{}
}
\toprule
Variant & PFM & CA & RG & CAD & IoU & mIoU \\
\midrule

VEOcc fusion
& \multicolumn{4}{c|}{Analytic fusion}
& 63.42 & 54.66 \\

No memory
& -- & -- & -- & $\checkmark$
& 61.33 & 52.03 \\

Mean update
& $\checkmark$ & -- & -- & $\checkmark$
& 66.25 & 57.49 \\

Gate only
& $\checkmark$ & -- & $\checkmark$ & $\checkmark$
& 66.86 & 58.16 \\

Candidate only
& $\checkmark$ & $\checkmark$ & -- & $\checkmark$
& 66.77 & 57.85 \\

Pointwise decode
& $\checkmark$ & $\checkmark$ & $\checkmark$ & --
& 66.51 & 57.81 \\

\midrule
Full
& $\checkmark$ & $\checkmark$ & $\checkmark$ & $\checkmark$
& \textbf{67.12} & \textbf{58.40} \\

\bottomrule
\end{tabular}
\end{minipage}
\hfill
%
\begin{minipage}[t]{0.495\linewidth}
\vspace{0pt}
\centering
\scriptsize

\caption{\textbf{Offline refinement ablation.}
PF: persistent feature; LF: local features;
LA: learned aggregation; GSR: global spatial refinement.}
\label{tab:offline_ablation}

\setlength{\tabcolsep}{2.0pt}

\begin{tabular}{
    @{}
    >{\raggedright\arraybackslash}p{0.29\linewidth}|
    C{0.065\linewidth}
    C{0.065\linewidth}
    C{0.065\linewidth}
    C{0.065\linewidth}|
    C{0.10\linewidth}
    C{0.11\linewidth}
    @{}
}
\toprule
Variant & PF & LF & LA & GSR & IoU & mIoU \\
\midrule

Online map
& -- & -- & -- & --
& 67.12 & 58.40 \\

Probability only
& -- & -- & -- & $\checkmark$
& 67.36 & 58.62 \\

Prob. + feature
& $\checkmark$ & -- & -- & $\checkmark$
& 67.54 & 59.03 \\

Prob. + local
& -- & $\checkmark$ & $\checkmark$ & $\checkmark$
& 67.56 & 58.94 \\

Mean fusion
& $\checkmark$ & $\checkmark$ & -- & $\checkmark$
& 67.59 & 59.05 \\

Pointwise refine
& $\checkmark$ & $\checkmark$ & $\checkmark$ & --
& 67.21 & 58.68 \\

\midrule
Full
& $\checkmark$ & $\checkmark$ & $\checkmark$ & $\checkmark$
& \textbf{67.79} & \textbf{59.37} \\

\bottomrule
\end{tabular}
\end{minipage}

\vspace{-10pt}
\end{table}

%% file: sections/5_conclusion.tex
\section{Conclusion}

We presented {\methodname}, a unified voxel-centric framework that connects
online incremental mapping and offline global map refinement through a
Persistent Voxel Feature Map. During exploration, reliability-guided feature
updates and context-aware decoding maintain an incrementally expanding global
map; after exploration, cached local evidence and spatial refinement further
improve the completed representation. {\methodname} establishes
state-of-the-art online mapping performance on EmbodiedOcc-ScanNet, while
offline refinement further improves the final map using the observations
already acquired during exploration. Zero-shot robotic experiments additionally
demonstrate transferability to unseen environments, onboard feasibility, and
floor-scale indoor mapping.
{\methodname} currently relies on externally provided camera poses, and its
persistent map and cached local evidence grow with the explored scene and
sequence length. Future work will investigate robustness to pose uncertainty
and more compact long-term scene representations.

%% file: sections/7_appendix.tex
\begin{center}
    {\Large\bfseries Appendix}
\end{center}
\medskip

\startcontents[appendix]
\section*{Appendix Contents}
\setcounter{tocdepth}{2}
\printcontents[appendix]{}{1}{}

\section{Additional Method Details}
\label{app:method_details}

\subsection{Local Predictor and Voxel Feature Extraction}
\label{app:local_details}

The local occupancy predictor follows VEOcc~\citep{veocc} without architectural
modification. Image features are extracted by a 2D backbone, while Depth
Anything V2~\citep{depthanythingv2} provides a monocular depth prior from the
same RGB input. The depth prior conditions the LSS-based lifting process,
projecting image features into the local voxel space without requiring an
external depth sensor. A 3D encoder then produces the voxel representation
$\mathbf{F}_t$, from which a lightweight occupancy head predicts frame-wise
semantic occupancy. Importantly, the subsequent mapping stages operate on
$\mathbf{F}_t$ before the final occupancy head, thereby preserving richer
feature information for cross-frame accumulation.

\subsection{Observation Alignment and Reliability Estimation}
\label{app:alignment_details}

The pose-conditioned adaptation in Sec.~\ref{subsec:online_mapping} is
implemented as a residual transformation conditioned on the normalized local
voxel coordinate and a 6D camera rotation encoding:
\begin{equation}
\widehat{\mathbf{f}}_{t,i}
=
\mathbf{f}_{t,i}
+
\phi\!\left(
[\mathbf{f}_{t,i},\mathbf{a}_{t,i},\mathbf{e}^{\mathrm{rot}}_t]
\right),
\end{equation}
where $\phi(\cdot)$ is a lightweight MLP. Its output layer is zero-initialized,
such that the module starts from an identity transformation.

For reliability estimation, we follow the Reliability-Aware Confidence
Modulation (RCM) of VEOcc~\citep{veocc}. For each local voxel observation $i$,
the confidence is computed from its camera-space depth and image-boundary
proximity:
\begin{equation}
c_{t,i}
=
\operatorname{clip}
\left(
\exp(-\alpha z_{t,i})
\exp(-\beta b_{t,i}),
c_{\min},1
\right),
\end{equation}
where $z_{t,i}$ denotes camera-space depth and $b_{t,i}$ measures proximity to
the image boundary. For the observations $\mathcal{G}_t(k)$ associated with
global voxel $k$, the voxel-wise reliability is computed as
\begin{equation}
r_t(k)
=
\frac{
\sum_{i\in\mathcal{G}_t(k)} c_{t,i}^{2}
}{
\sum_{i\in\mathcal{G}_t(k)} c_{t,i}
}.
\end{equation}
This reliability is computed independently from the aligned voxel feature
$\mathbf{x}_t(k)$ and is used to condition the subsequent feature update.

\subsection{Persistent Map Update and Online Decoding}
\label{app:memory_details}

For each mapped voxel, we additionally maintain the historical reliability
$\bar r_t(k)$ and observation count $n_t(k)$. For a repeated observation, they
are updated as
\begin{equation}
\bar r_t(k)
=
\frac{
n_{t-1}(k)\bar r_{t-1}(k)+r_t(k)
}{
n_{t-1}(k)+1
},
\qquad
n_t(k)=n_{t-1}(k)+1.
\end{equation}
The observation count is used only to maintain the running reliability and is
not provided to the learned updater.

For a re-observed voxel, we use the shorthand
$\mathbf{h}=\mathbf{h}_{t-1}(k)$ and
$\mathbf{x}=\mathbf{x}_t(k)$. The standardized feature distance is
\begin{equation}
D(\mathbf{h},\mathbf{x})
=
\sqrt{
\operatorname{mean}
\left[
(\mathbf{h}^{n}-\mathbf{x}^{n})^2
\right]
+\epsilon
},
\end{equation}
where $\mathbf{h}^{n}$ and $\mathbf{x}^{n}$ denote the standardized historical
and current features.

We characterize the relation between the historical and current features as
\begin{equation}
\label{eq:feature_relation}
\mathbf{f}^{\mathrm{rel}}_t(k)
=
\left[
\mathbf{h},
\mathbf{x},
\mathbf{h}-\mathbf{x},
\mathbf{h}\odot\mathbf{x},
\operatorname{cos}(\mathbf{h},\mathbf{x}),
D(\mathbf{h},\mathbf{x})
\right].
\end{equation}
Together with semantic uncertainty $E_t(k)$ derived from the local prediction,
the current reliability $r_t(k)$, and the historical average reliability
$\bar r_{t-1}(k)$, the complete update descriptor is
\begin{equation}
\label{eq:update_descriptor}
\boldsymbol{\xi}_t(k)
=
\left[
\mathbf{f}^{\mathrm{rel}}_t(k),
E_t(k),
r_t(k),
\bar r_{t-1}(k)
\right].
\end{equation}

The descriptor jointly conditions the candidate adaptation and
reliability-guided gating branches. The two branches share a lightweight
feature encoder, and their output layers are zero-initialized such that the
initial update reduces to reliability-ratio interpolation.

The online decoder is implemented by a submanifold sparse convolution followed
by a pointwise sparse classifier.

\section{Training Details}
\label{app:training_details}

{\methodname} is trained in three sequential stages, with modules optimized in
earlier stages frozen during subsequent training. All stages are trained on
four NVIDIA RTX 3090 GPUs. Table~\ref{tab:training_details} summarizes the
batch configuration, training schedule, and training time for each stage.

\input{tables/train_detail}
\input{tables/local_lr_ablation}

\subsection{Local Predictor Training}
\label{app:training_local}

We train the local predictor for 10 epochs using AdamW with a base learning
rate of $2\times10^{-3}$ and a $0.1$ learning-rate multiplier for the image
backbone. Following VEOcc~\citep{veocc}, the local training objective is
\begin{equation}
\mathcal{L}_{\mathrm{local}}
=
100\mathcal{L}_{\mathrm{focal}}
+
\mathcal{L}_{\mathrm{Lovasz}}
+
\mathcal{L}_{\mathrm{sem}}
+
\mathcal{L}_{\mathrm{geo}}.
\end{equation}
The monocular depth prior is not additionally supervised during this stage.

\subsection{Online Incremental Mapping}
\label{app:training_online}

We freeze the local predictor and optimize the pose-conditioned feature
adaptation, Reliability-Guided Incremental Update, and Context-Aware Decoding
modules for 3 epochs. AdamW is used with a learning rate of
$2\times10^{-4}$ and a linear warm-up over 603 iterations. The online training
objective is
\begin{equation}
\mathcal{L}_{\mathrm{online}}
=
\mathcal{L}_{\mathrm{CE}}
+
\mathcal{L}_{\mathrm{Lovasz}}
+
\mathcal{L}_{\mathrm{sem}}
+
\mathcal{L}_{\mathrm{geo}},
\end{equation}
where the cross-entropy term uses inverse-log class-frequency weighting.

Training follows the causal mapping process. The persistent map state is
propagated across frames within each trajectory, while gradients through the
historical state are detached after each optimization step. Therefore, the
model is optimized with single-step truncated temporal back-propagation rather
than back-propagating through the complete sequence.
\input{figures/realworld_scaling}

\subsection{Offline Global Map Refinement}
\label{app:training_offline}

Finally, we freeze both the local predictor and online incremental mapping
modules and jointly optimize Local Evidence Aggregation and Global Spatial
Refinement for 30 epochs. AdamW is used with a peak learning rate of
$10^{-3}$, one epoch of warm-up, and cosine decay to $10^{-6}$.
The offline refinement stage uses the same combination of cross-entropy,
Lovasz, semantic-scaling, and geometric-scaling losses as the online stage.

\section{Additional Experimental Analysis}
\label{app:additional_experiments}

\subsection{Learning-Rate Analysis for Local Prediction}
\label{app:local_lr}

We further analyze the effect of the learning rate on the unchanged VEOcc
local predictor. As shown in Table~\ref{tab:local_lr_ablation}, increasing the
learning rate consistently improves local occupancy prediction, indicating
that the original architecture retains additional optimization potential.
Based on this analysis, we adopt the selected learning-rate setting for all
subsequent experiments while keeping the local prediction architecture
unchanged.

\input{figures/realworld_online_compare}

\subsection{Online Scaling in Real-World Deployment}
\label{app:deployment_scaling}

Fig.~\ref{fig:realworld_scaling} provides a finer-grained view of the online
deployment behavior on the Laboratory Floor sequence. As exploration proceeds,
the Persistent Voxel Feature Map grows to nearly one million voxels, while
process memory increases gradually with the accumulated scene state. Online
latency also rises with map growth but remains below the 1\,s reference for
all processed keyframes in this sequence. These results complement the
aggregate statistics in Table~\ref{tab:realworld} by showing how runtime and
memory evolve throughout long-horizon floor-scale mapping.



\subsection{Real-World Online Mapping Comparison}
\label{app:realworld_online_comparison}

Fig.~\ref{fig:realworld_online_compare} presents additional qualitative comparison of
online incremental mapping in real-world robotic scenes. The first two rows
show self-collected sequences previously used in VEOcc, and the last two rows
show the two zero-shot deployment sequences reported in the main paper of
VEOcc++. In all four cases, VEOcc++ produces cleaner and more spatially
consistent occupancy maps. The highlighted local regions further show improved
continuity of elongated structures, more complete recovery of cabinets and
walls, and stronger semantic consistency for furniture and small objects.
These results provide additional evidence that the proposed feature-level
mapping pipeline improves real-world online occupancy mapping beyond analytic
prediction-level fusion.

\stopcontents[appendix]

%% file: tables/train_detail.tex
\begin{table}[t]
\centering
\captionsetup{font=small}
\caption{
\textbf{Training configuration and time of the three stages.}
}
\label{tab:training_details}
\footnotesize
\setlength{\tabcolsep}{5pt}
\begin{tabular}{lcccc}
\toprule
Stage
& Per-GPU Batch
& Total Batch
& Epochs
& Training Time (h) \\
\midrule
Local Predictor
& 4 images
& 16 images
& 10
& 24.8 \\

Online Incremental Mapping
& 1 trajectory (30 frames)
& 4 trajectories
& 3
& 1.3 \\

Offline Global Map Refinement
& 1 scene map
& 4 scene maps
& 30
& 1.9 \\
\bottomrule
\end{tabular}
\end{table}

%% file: tables/local_lr_ablation.tex
\begin{table}[t]
    \centering
    \captionsetup{font=small}
    \caption{\textbf{Local learning-rate study on Occ-ScanNet.}
    All results are measured at epoch 10 using the same architecture
    and remaining training configuration.
    The best results are marked in \textbf{bold}.}
    \label{tab:local_lr_ablation}
    \small
    \setlength{\tabcolsep}{10pt}
    \begin{tabular}{lccc}
        \toprule
        Learning rate & Multiplier & IoU & mIoU \\
        \midrule
        $4\times10^{-4}$ (VEOcc; \citealp{veocc}) & $1\times$ & 64.55 & 55.49 \\
        $8\times10^{-4}$            & $2\times$ & 66.11 & 57.35 \\
        $1.6\times10^{-3}$          & $4\times$ & 66.90 & 58.08 \\
        $2\times10^{-3}$ (VEOcc++, Ours) & $5\times$ &
        \textbf{66.98} & \textbf{58.17} \\
        \bottomrule
    \end{tabular}
\vspace{-10pt}
\end{table}

%% file: figures/realworld_scaling.tex
\begin{figure}[t]
    \centering
    \captionsetup{font={small}}
    \includegraphics[width=0.8\linewidth]{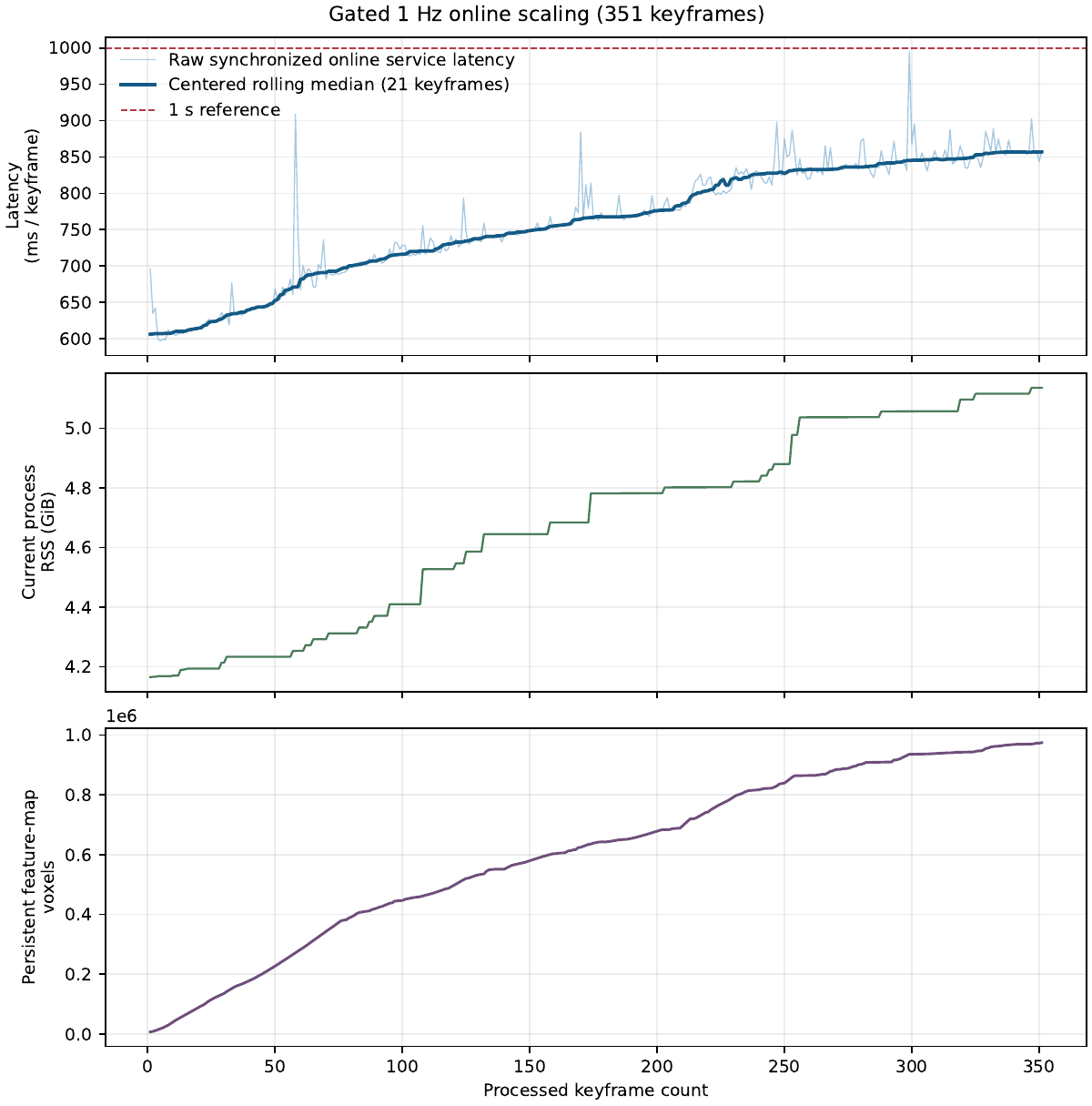}
    \caption{
    \textbf{Online deployment scaling on the Laboratory Floor sequence.}
    We track runtime and memory as the global map grows over 351 motion-gated
    keyframes. The panels show per-keyframe online service latency, current
    process RSS, and Persistent Voxel Feature Map size, respectively. The
    centered rolling median over 21 keyframes highlights the latency trend,
    while the dashed line indicates the 1\,s reference.
    }
    \label{fig:realworld_scaling}
\end{figure}

%% file: figures/realworld_online_compare.tex
\begin{figure}[t]
    \centering
    \captionsetup{font={small}}
    \includegraphics[width=\linewidth]{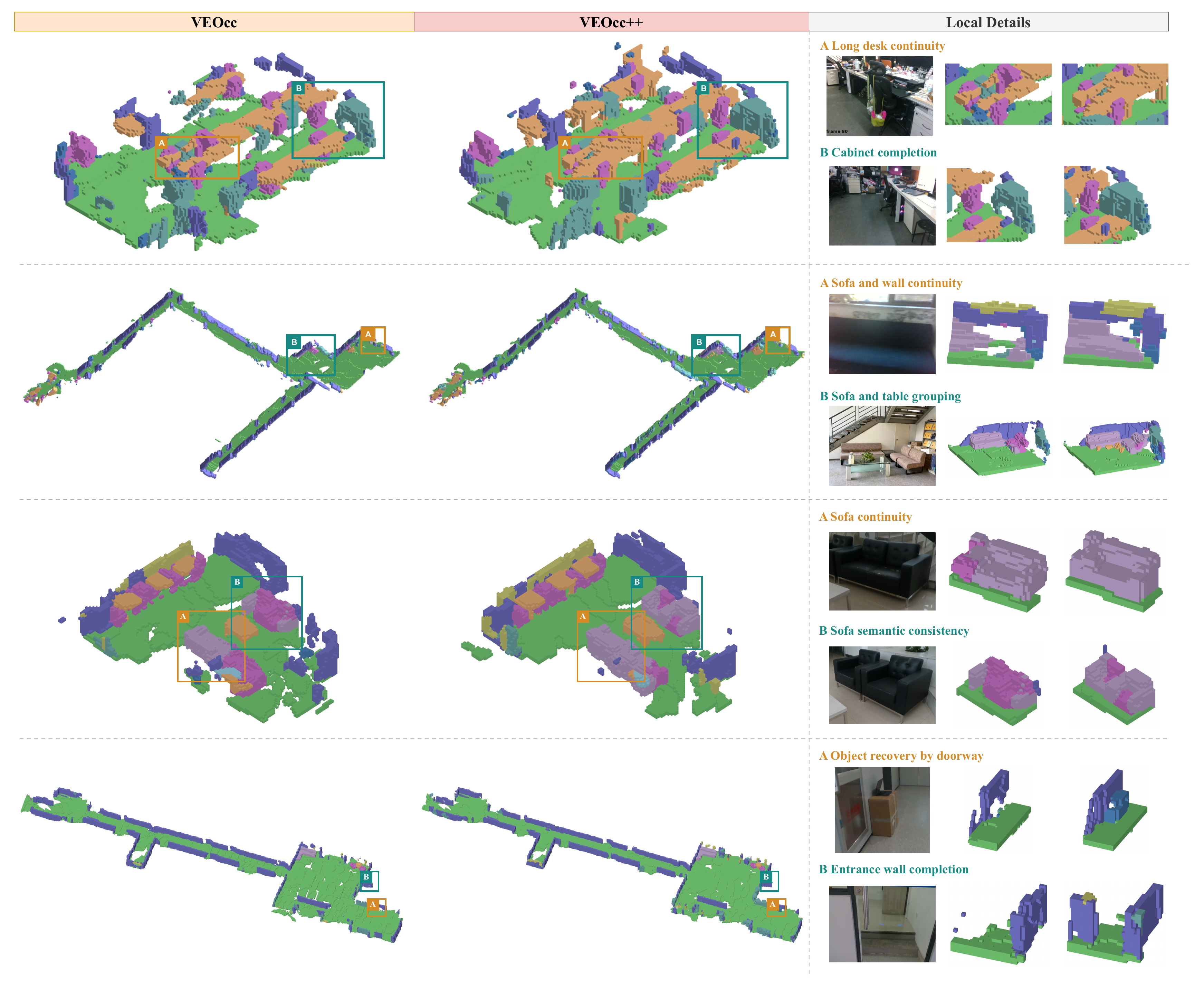}
    \caption{
    \textbf{Qualitative comparison of online incremental mapping on real-world robotic sequences.}
    The first two rows show self-collected sequences previously used in VEOcc,
    while the last two rows correspond to the two zero-shot deployment sequences
    reported in the main paper of VEOcc++. For each sequence, we compare the
    online maps produced by VEOcc and VEOcc++, and show representative local
    regions in the right column. VEOcc++ yields more complete structures, better
    continuity of extended objects and surfaces, and improved semantic consistency
    in the highlighted regions.
    }
    \label{fig:realworld_online_compare}
    \vspace{-10pt}
\end{figure}